\documentclass[10pt,a4paper]{article}
\usepackage[utf8]{inputenc}
\usepackage{amsmath}
\usepackage{amsfonts}
\usepackage{amssymb}
\usepackage{graphicx}

\title{State-of-the-Art in Retinal Optical Coherence Tomography Image Analysis}

\author{Ahmadreza Baghaie, Roshan M. D'souza, Zeyun Yu \\ \textit{University of Wisconsin-Milwaukee, WI, USA}}

\begin{document}
\maketitle

\begin{abstract}
Optical Coherence Tomography (OCT) is one of the most emerging imaging modalities that has been used widely in the field of biomedical imaging. From its emergence in 1990's, plenty of hardware and software improvements have been made. Its applications range from ophthalmology to dermatology to coronary imaging etc. Here, the focus is on applications of OCT in ophthalmology and retinal imaging. OCT is able to non-invasively produce cross-sectional volume images of the tissues which are further used for analysis of the tissue structure and its properties. Due to the underlying physics, OCT images usually suffer from a granular pattern, called \textit{speckle} noise, which restricts the process of interpretation, hence requiring specialized noise reduction techniques to remove the noise while preserving image details. Also, given the fact that OCT images are in the $\mu m$ -level, further analysis in needed to distinguish between the different structures in the imaged volume. Therefore the use of different segmentation techniques are of high importance. The movement of the tissue under imaging or the progression of disease in the tissue also imposes further implications both on the quality and the proper interpretation of the acquired images. Thus, use of image registration techniques can be very helpful. In this work, an overview of such image analysis techniques will be given.
\end{abstract}

\section{Introduction}
Optical Coherence Tomography (OCT) is a powerful imaging system for acquiring 3D volume images of tissues non-invasively. In simple terms, OCT can be considered as echography with light \cite{Pizurica 2008}. Unlike the echography which is done by sound waves, OCT imaging is not time-of-flight based and instead produces the image based on the interference patterns. Fig. 1 shows a typical retinal OCT image with false color. Throughout the past 2 decades, many improvements have been achieved regarding the OCT imaging system which not only improved the acquisition time but also the quality of the acquired images. Nowadays taking $ \mu m$-level volume images of the tissues is very common especially in ophthalmology and retinal imaging. Therefore the need for specialized OCT image analysis techniques is of high interest, making this by far the most attractive area in biomedical imaging \cite{Tomlins 2005}.

\begin{figure}
\begin{center}
\includegraphics[scale=.3]{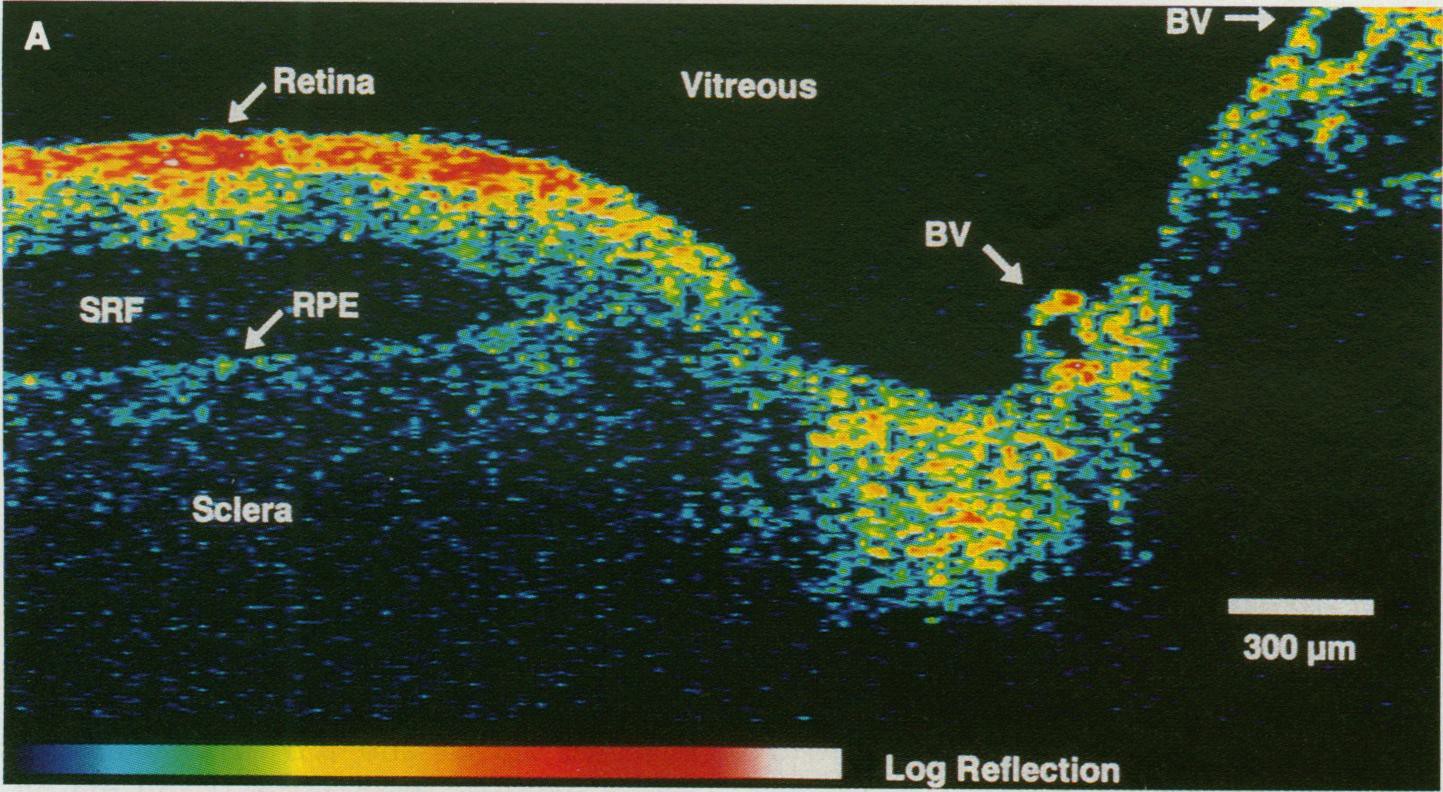}
\caption{Optical coherence tomograph of human retina and optic nerve \cite{Huang 1991}}
\end{center}
\end{figure}

From a practical point of view, several aspects of image processing techniques are of high demand in analyzing OCT images which will be discussed in more detail in this article. As a pre-processing step, noise reduction techniques are usually applied to OCT images. Due to the physical implications of coherent imaging, OCT images usually suffer from granular patterns which block fine details in the tissue images. Plenty of techniques are proposed to remedy this issue some of which will be mentioned in the following sections. 

Use of image segmentation methods is another area which is widely explored in OCT image analysis, especially in ophthalmology. For example, layered-structure of retina has been the subject of plenty of papers in the past few years. The methods that are used are usually specialized versions of the corresponding methods in general image segmentation, ranging from graph-based techniques to active contour methods etc. They will be further discussed later in this article.

Image registration also has its own place among the many image processing techniques used in OCT image analysis. This becomes more obvious when one realizes that OCT imaging is a $\mu m$-level imaging system which means even a small movement in the subject can have a big impact on the result. Even though the OCT imaging systems are very fast, still, the volume to be imaged is very dense so having an informative and comprehensive volume image requires hundreds of cross-sections which makes the imaging modality vulnerable to small movements. Also with the progress of degeneration in the tissue due to illness, image registration can be very useful in tracking the changes over time.

Based on the above-mentioned notes, the article is organized as follows: In Section 2 an overview of OCT imaging and different techniques that are used to acquire and reconstruct the images is given. Section 3 focuses on noise reduction techniques. In Section 4 a few techniques for OCT image segmentation, with a focus on retinal layer segmentation are reviewed. Section 5 contains an overview of the use of image registration techniques in OCT image analysis. Section 6 concludes the article with pointers on some of the future paths that can be taken for further investigations. 

\section{Optical Coherence Tomography (OCT) Imaging}
OCT imaging works based on the interference between a split and later re-combined broadband optical field \cite{Tomlins 2005}. Fig. 1 displays a typical OCT imaging system. The light beam from the source is exposed to a beam splitter and travels in two paths: one toward a moving reference mirror and the other to the sample to be imaged. The reflected light from both the reference mirror and the sample will be fed to a photo detector in order to observe the interference pattern. The sample usually contains particles (or layers) with different refractive indexes and the variation between neighbor refractive indexes causes intensity peaks in the interference pattern detected by the photo detector. By translating the reference mirror, a time domain interference pattern can be obtained. Also, by frequency domain measurements of the output spectrum, depth information can be derived. The result will be a line scan (aka A-scan) of depth of the 3D volume to be imaged. Using multiple A-scans, 2D cross-sections and 3D volumes can be constructed.

\begin{figure}
\begin{center}
\includegraphics[scale=.4]{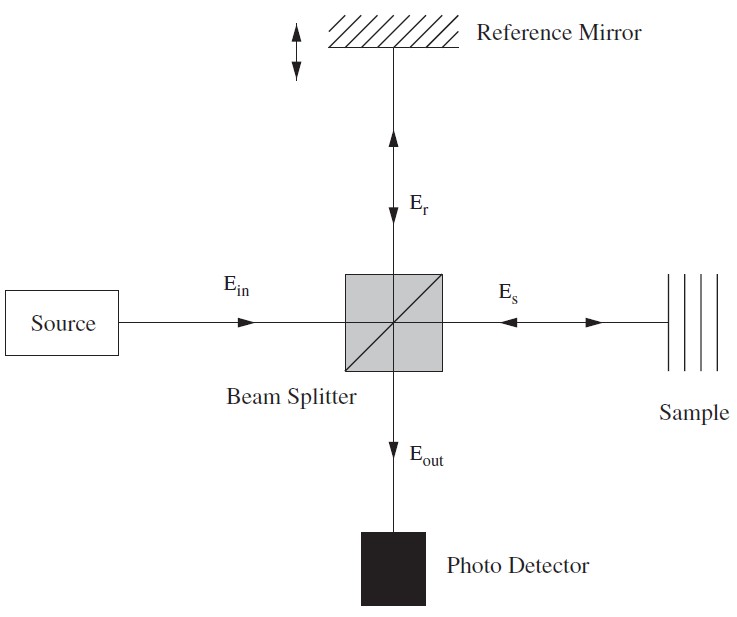}
\caption{A typical OCT system \cite{Tomlins 2005}}
\end{center}
\end{figure}

The most popular OCT imaging system is called time-domain OCT (TD-OCT) in which a reference mirror is translated to match the optical path from reflections within the sample \cite{Huang 1991}. Unlike the TD-OCT, in Fourier-domain OCT (FD-OCT) there is no need for moving parts in the design of the imaging system to obtain the axial scans \cite{Fercher 1995}. In FD-OCT, the reference path length is fixed and the detection system is replaced with a spectrometer. Because of not having any moving parts, the speed of FD-OCT systems in acquiring images is very high in comparison to TD-OCT.

Moving away from the classical behavior of light and going toward the quantum nature of light, Quantum OCT (Q-OCT) is a new imaging modality \cite{Nasr 2003}. Full field OCT (FF-OCT) is another optical coherence imaging technique in which a CCD camera is placed at the output instead of the single detector of the TD-OCT to capture 2D \textit{en face} image in a single exposure \cite{Beaurepaire 1998}. Taking into account the polarization state of the light, polarization-sensitive OCT (PS-OCT) is another technique for imaging the birefringence within a biological sample \cite{Boer 2002}. Doppler OCT (D-OCT) which is also called optical Doppler tomography (ODT) is a combination of OCT imaging system with laser Doppler flowmetry. This system allows for the quantitative imaging of fluid flow in a highly scattering medium; such as monitoring \textit{in vivo} blood flow beneath the skin \cite{Chen 1998}.

In the following sections more focus will be given to the software based image analysis techniques rather than hardware of OCT systems. 

\section{Noise Reduction}
Speckle is a fundamental property of the signals and images acquired by narrow-band detection systems like SAR, ultrasound and OCT. In OCT, not only the optical properties of the system, but also the motion of the subject to be imaged, size and temporal coherence of the light source, multiple scattering, phase deviation of the beam and aperture of the detector can affect the speckle \cite{Schmitt 1999}. Two main processes affect the spatial coherence of the returning light beam which is used for image reconstruction: 1) multiple back-scattering of the beam, and 2) random delays for the forward-propagating and returning beam caused by multiple forward scattering. In the case of tissue imaging, since the tissue is packed with sub-wavelength diameter particles which act as scatterers, both of these phenomena contribute to the creation of speckle. As stated in \cite{Schmitt 1999}, two types of speckle are present in OCT images: signal-carrying speckle which originates from the sample volume in the focal zone; and signal-degrading speckle which is created by multiple-scattered out-of-focus light.  The latter kind is what that is considered as speckle noise. Fig. 3 displays the common scene in retinal OCT imaging: a highly noisy image.

\begin{figure}
\begin{center}
\includegraphics[scale=.2]{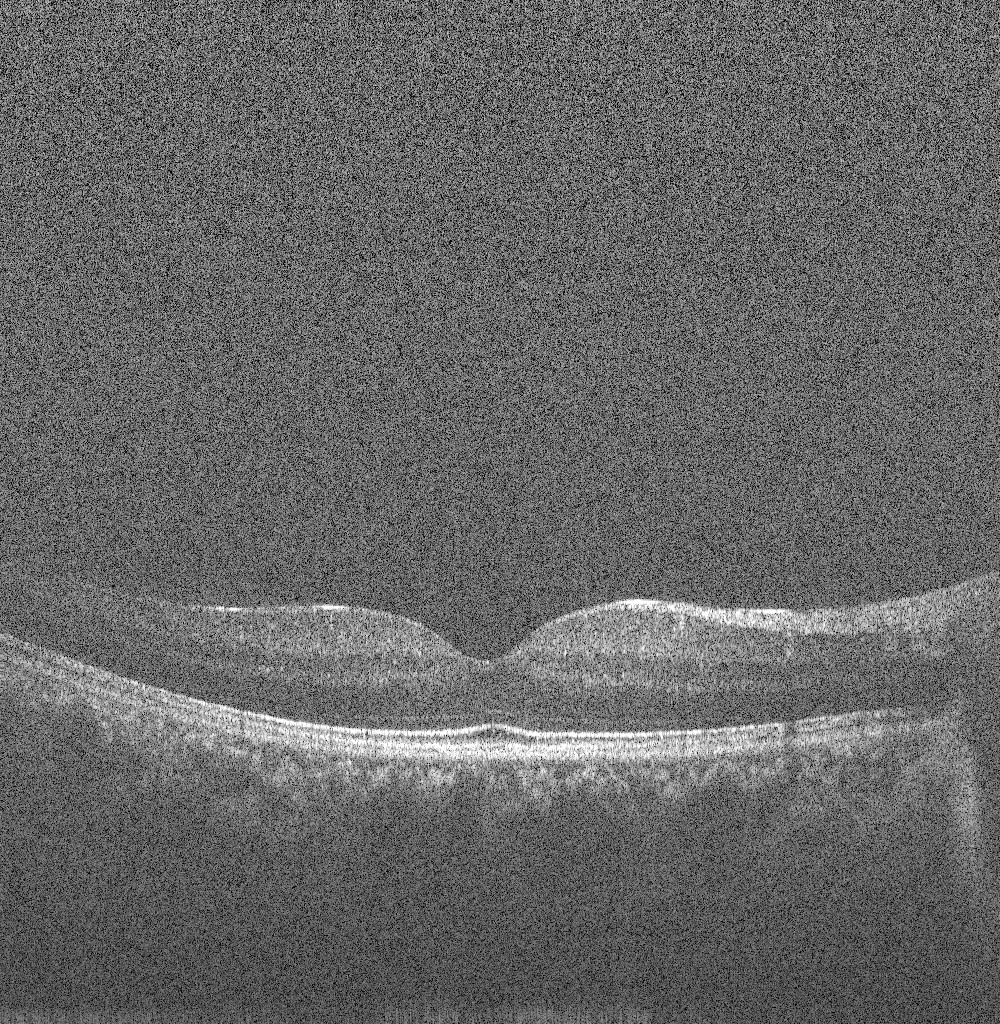}
\caption{Typical retinal OCT image degraded by speckle noise}
\end{center}
\end{figure}

The distribution of the speckle can be represented with a Rayleigh distribution. Speckle is considered as a multiplicative noise, in contrast to Gaussian additive noise. Due to the limited dynamic range of displays, OCT signals are usually compressed by a logarithmic operator applied to the intensity information. After this, the multiplicative speckle noise is transformed to additive noise and can be further treated \cite{Salinas 2007}. 

OCT noise reduction techniques can be divided into two major classes: 1) methods of noise reduction during the acquisition time and 2) post-processing techniques. In the first class, aka compounding techniques, multiple uncorrelated recordings are averaged. Among them, spatial compounding \cite{Avanaki 2013}, angular compounding \cite{Schmitt 1997}, polarization compounding \cite{Kobayashi 1991} and frequency compounding \cite{Pircher 2003} techniques can be mentioned. 

Usually the methods from the first class are not preferred since they require multiple scanning of the same data which extends the acquisition time. Plus, these techniques can be very restrictive in terms of different OCT imaging systems in use. Therefore the use of more general post-processing techniques is more favorable. 

In the literature, plenty of post-processing methods are proposed for speckle noise reduction of OCT images. There are a few classic de-speckling techniques for noise reduction that are successfully used in OCT images too: Lee \cite{Lee 1980}, Kuan \cite{Kuan 1985} and Frost \cite{Frost 1982} filters.

One of the very interesting groups of methods for speckle reduction use the well-known anisotropic diffusion method \cite{Perona 1990}. Due to its poor performance in the case of very noisy images, there are different variations of this method in the literature \cite{Yu 2002, Abd 2002, Krissian 2007}. In \cite{Salinas 2007} a complete comparison is provided for regular anisotropic diffusion and complex anisotropic diffusion approaches for denoising of OCT images. Another example of using anisotropic diffusion is in \cite{Puvanathasan 2009} where an Interval type-II fuzzy approach is used for determining the diffusivity function of the anisotropic diffusion. 

Another important group of widely used methods for de-speckling of OCT images take advantage of multiscale/multiresolution geometric representation techniques \cite{Pizurica 2008, Jacques 2011}. The main reason for this is that these representations of the images compress the essential information of the image into a few large coefficients while noise is spread among all of the coefficients. Generally, for such methods a transform domain that provides a sparse representation of images and an optimal threshold is needed. Using the optimal threshold, hard- or soft-thresholding can be done for coefficient shrinkage in order to reduce the noise. This threshold can be a constant or even better, sub-band adaptive or spatially adaptive. For example, in OCT images, since it is known that most of the features are horizontal, higher thresholds can be applied to the vertical coefficients.  Wavelet transform is one example of such methods which has been widely used in this area \cite{Adler 2004}. Even though wavelet transform has shown promising results, its lack of directionality imposes some limitations in properly denoising OCT images. This can be further improved by use of dual-tree complex wavelet transform (DT-CWT) which doubles the directional information of wavelet transform \cite{Selesnick 2005}. Still, there are better transform domains which offer more proper representations for OCT images. In the past few years use of Curvelet transform \cite{Jian 2009, Jian 2010}, Contourlet transform \cite{Guo 2013, Xu 2013} and Ripplet transform \cite{Gupta 2014} showed promising results in the denoising of speckle noise. Fig. 4 shows the result of curvelet coefficients' shrinkage for speckle noise reduction.

\begin{figure}
\begin{center}
\includegraphics[scale=.35]{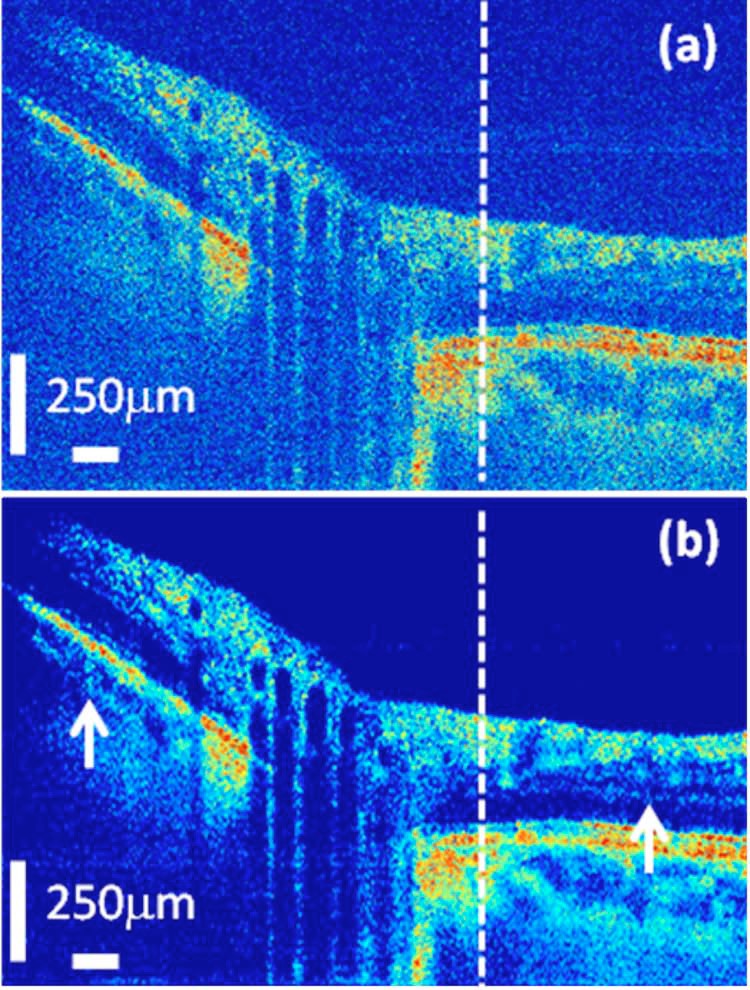}
\caption{FD-OCT image of optical nerve head, before (a) and after (b) curvelet coefficients' shrinkage-based speckle noise reduction \cite{Jian 2009}}
\end{center}
\end{figure}

Compressive sensing and sparse representation are new directions that are taken for image acquisition and noise reduction. Very recent examples are \cite{Fang 2012, Fang 2013} which use dictionary learning and sparse representation for noise reduction and image reconstruction of OCT. The process for these methods usually involves dividing the images into overlapping patches and training dictionaries. This is usually done by taking patches from both high-SNR-high-resolution (HH) and low-SNR-low-resolution (LL) samples from a volume data. The trained dictionary is then used in the process of denoising or reconstruction of images which acquired using random and incomplete patterns of data reading.

\section{Image Segmentation}
Image segmentation is a very active area in the field of medical image analysis. The most difficult step in any medical image analysis system is the automated localization and extraction of the structures of interest \cite{DeBuc 2011}. A critical role for medical image segmentation in OCT images involves the segmentation and extraction of regions of the image in order to quantify areas and volumes of interest in imaged tissues for further analysis and diagnosis. 

The signal strength in OCT images raises from the intrinsic differences in tissue optical properties. In retina, multiple layers of different types of cells can be seen in OCT images. Fig. 2 displays different views of retina \cite{Garvin 2008}.

\begin{figure}
\begin{center}
\includegraphics[scale=.2]{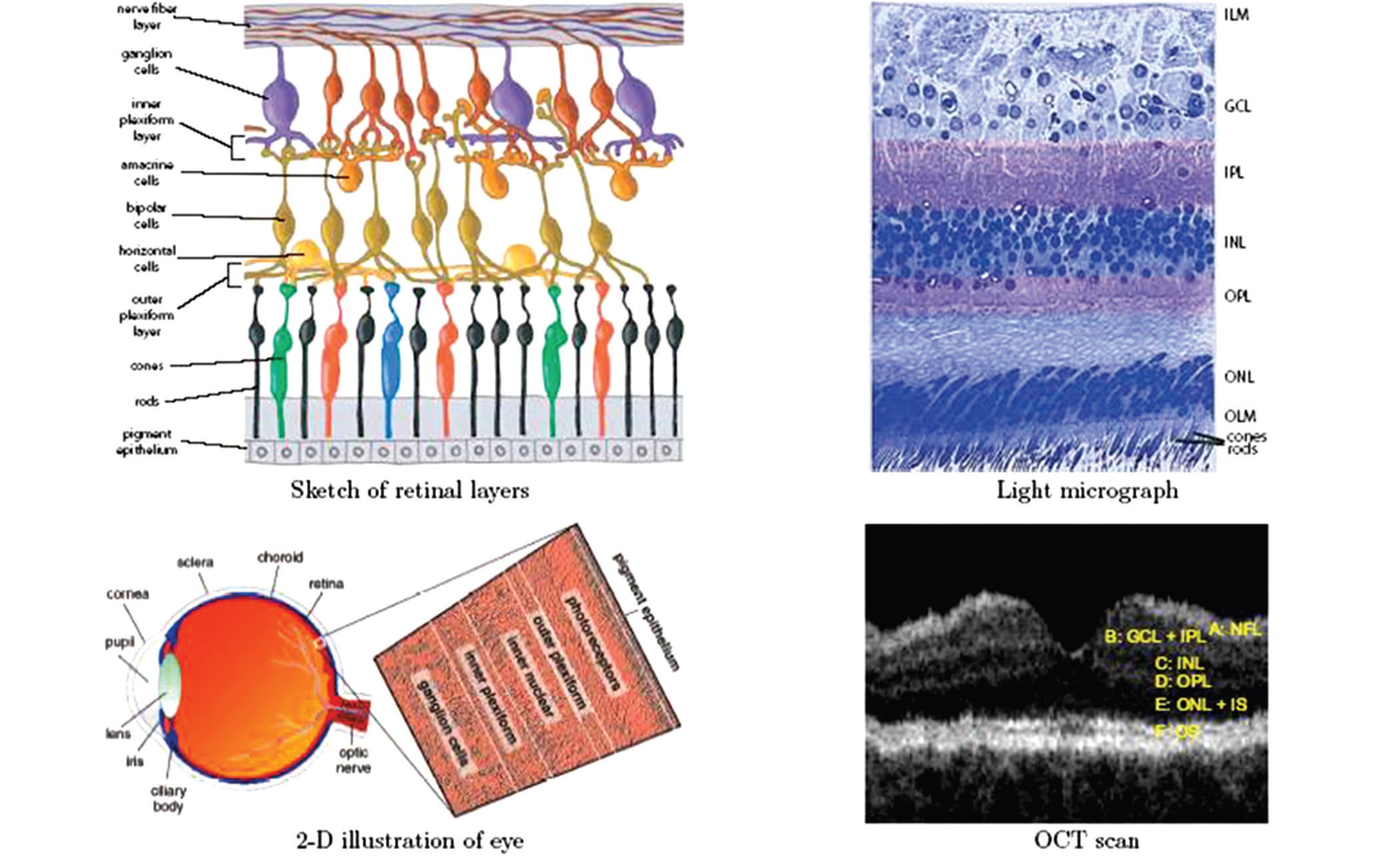}
\caption{Different views of the retina (a) Schematic illustration of cellular layers of retina, (b) Light micrograph of a vertical scan through central human retina, (c) Cross section of the eye with illustration of the retina, (d) OCT view of macular retina \cite{Garvin 2008}}
\end{center}
\end{figure}

During the last two decades plenty of methods are proposed for segmentation of OCT images. For a more complete review on the methods the reader is referred to \cite{DeBuc 2011} and \cite{Kafieh 2013}. Here an overview of the different classes of methods will be given.

Four major problems are encountered in the segmentation of OCT images:
\begin{enumerate}
\item The presence of speckle noise complicates the process of the precise identification of the retinal layers, therefore most of the segmentation methods require pre-processing steps to reduce the noise.
\item Intensity of homogeneous areas decreases with the increase of the depth of imaging. This is due to the fact that the intensity pattern in OCT images is the result of the absorption and scattering of light in the tissue. 
\item Optical shadows imposed by blood vessels also can affect the performance of segmentation methods.
\item Quality of OCT images degrades as a result of motion artifacts or sub-optimal imaging conditions.
\end{enumerate}

Methods of retinal segmentation can be categorized in five classes: 1) Methods applicable to A-scans, 2) intensity-based B-scans analysis, 3) active contour approaches, 4) analysis methods using artificial intelligence and pattern recognition techniques and 5) segmentation methods using 2D/3D graphs constructed from the 2D/3D OCT images \cite{Kafieh 2013, Kafieh 2013b}. 

A-scan based methods  consider the difference in the intensity levels to extract the edge information. Due to the effects of speckle noise, these methods are usually used for determining the most significant edges on the retina. Examples of such methods are presented in \cite{Huang 1991, Hee 1995, Dillenseger 2000, Koozekanani 2001, Shahidi 2005}. A-scan based methods lack the contribution from 2D/3D data while the computational time is hight and the accuracy is low. Taking advantage of better denoising methods, intensity-based B-scan analysis approaches work based on the intensity variations and gradients which are still too sensitive and make these methods case-dependent \cite{Baroni 2007, Tao 2008, Kajic 2010}.

Active contour methods also provide promising results for segmentation of retinal layers in OCT images. Methods presented in \cite{Fernandez 2004}, \cite{Mishra 2009} and \cite{Yazdanpanah 2011} are three well-known examples. Fig. 6 shows the results of layer segmentation using the active contour-based method proposed in \cite{Yazdanpanah 2011} for three OCT images.

\begin{figure}
\begin{center}
\includegraphics[scale=.4]{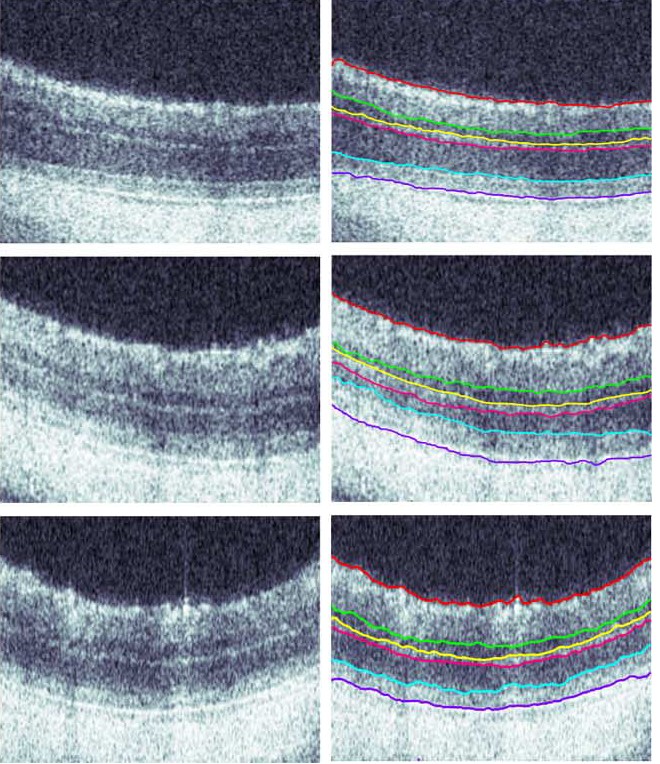}
\caption{Results of segmentation for three different retinal OCT images with active contour method proposed in \cite{Yazdanpanah 2011}}
\end{center}
\end{figure}

Use of pattern recognition based methods is a new direction that attracted researchers for more explorations. To name a few, in \cite{Fuller 2007} a support vector machine (SVM) is used to perfom the semi-automatic segmentation of retinal layers. In \cite{Meyer 2008} Fuzzy C-means clustering method is used for layer segmentation. The use of random forest classifier for segmentation is also reported in \cite{Lang 2013}.

2D/3D graph based methods are probably the best among all of the segmentation methods. During the past few years several variations of this class are introduced in the literature. Examples of such methods can be found in \cite{ Kafieh 2013b, Garvin 2008b, Chiu 2010, Chiu 2012, Srinivasan 2014}. The method in \cite{Kafieh 2013b} is based on diffusion maps \cite{Coifman 2006} and takes advantage of regional image texture rather than edge information. Therefore it is more robust in low contrast and poor layer-to-layer image gradients. On the other hand methods presented in \cite{Chiu 2010, Chiu 2012, Srinivasan 2014} use the graph-cuts and shortest path method for segmentation of retinal layers. Fig. 7 shows the final 3D visualization of the segmented layers using the diffusion -based method proposed in \cite{Kafieh 2013b}.

\begin{figure}
\begin{center}
\includegraphics[scale=.25]{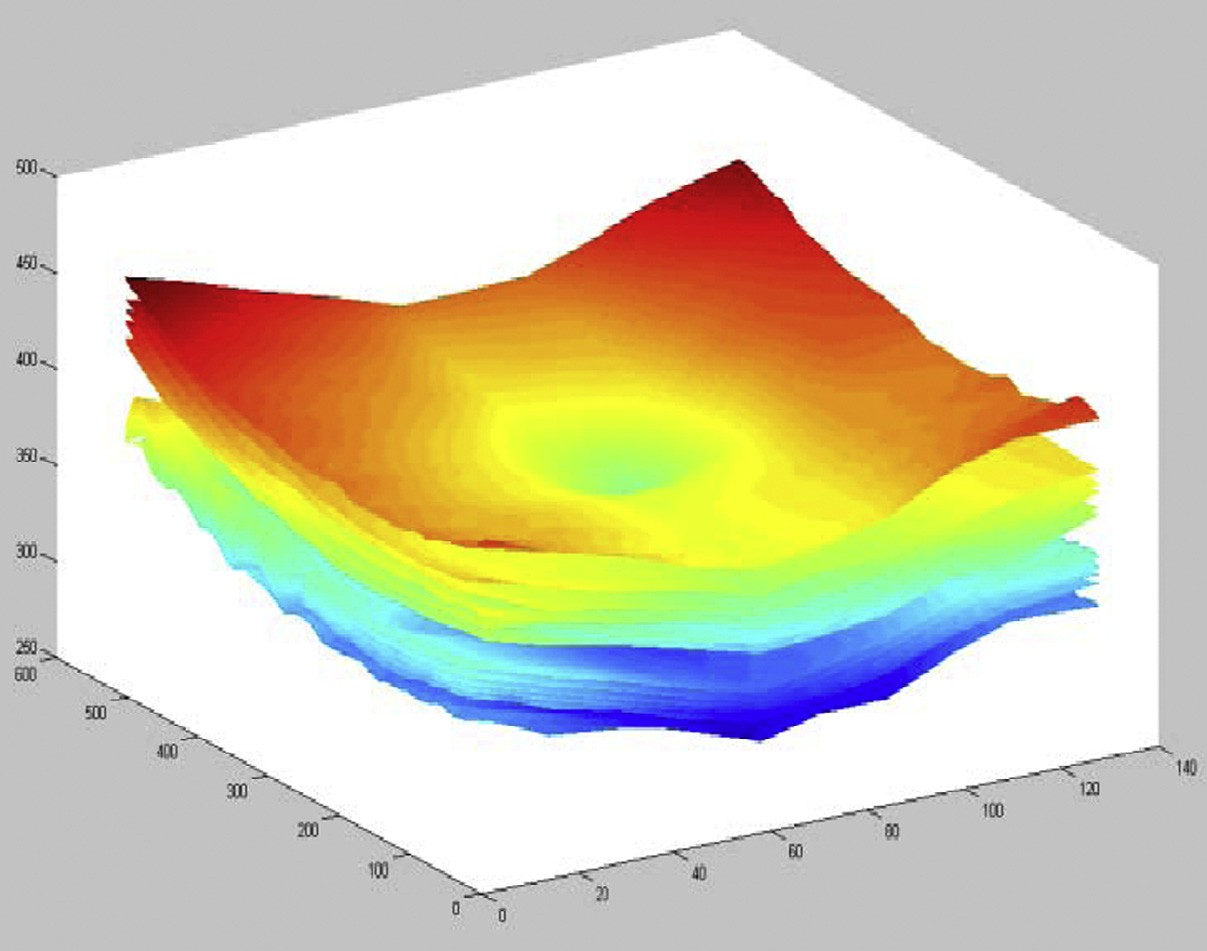}
\caption{3D visualization of the final result of segmentation \cite{Kafieh 2013b}}
\end{center}
\end{figure}

Despite the plethora of different methods in this area, retinal layer segmentation is still an open problem with lots of room for improvement. Here, only a few of the well-known methods are shortly introduced and the reader is referred to the mentioned articles and references therein.

\section{Image Registration}

Having two input images, template image and reference image, image registration tries to find a valid optimal spatial transformation to be applied to the template image to make it more similar to the reference image \cite{Zitova 2003}. Therefore the process of image registration consists of an optimization process fulfilling some imposed constraints. The applications of medical image registration range from mosaicing of retinal images \cite{Patton 2006} to slice interpolation \cite{Baghaie 2014} etc. 

There are two different classes of image registration techniques: 1) Parametric approaches and 2) Non-parametric approaches \cite{Modersitzki 2003}. In general, the process of finding the valid optimal spatial transformation can be formulated as $ E[u]=D[R, T\circ u]+ \alpha S $ in which $ R $ and $ T $ are reference and template images respectively, $ u $ is the displacement field, $ D $ is the similarity/dissimilarity measure (Mutual Information (MI), Normalized MI (NMI), Cross-Correlation (CC), Normalized CC, Sum of Squared Differences (SSD) and etc), $ S $ is the regularization term (which imposes additional constraints on the deformation), $ \alpha $ is the coefficient that determines the amount of regularization and $ E $ is the energy functional which depending on the problem either should be minimized or maximized \cite{Sotiras 2013}.

There are several different applications for using image registration approaches in OCT image analysis. One of the basic applications is in speckle noise reduction. As mentioned in Section 3, hardware-based noise reduction techniques take the average of several uncorrelated scans in order to reduce the noise. The same thing can happen after finishing image acquisition too. The main idea is to gather several images of the same cross-section in retina, register them together and take the average. Having a small movement in the eye provides us with an uncorrelated pattern of speckle. Having $ N $ B-scans, the SNR can be improved by a factor of $ \sqrt{N} $. In \cite{Jorgensen 2007} a dynamic programming based method is used for compensation of the movements between several B-scans and reducing the speckle noise. A hierarchical model-based motion estimation scheme based on an affine-motion model is used in \cite{Alonso-Caneiro 2011} for registering multiple B-scans to be used for speckle reduction. Fig. 6 shows the result of affine registering the images from close locations in retina and averaging. Recently, the use of low-rank/sparse decomposition based batch alignment has been investigated for speckle noise reduction in OCT images too. Taking advantage of Robust Principal Component Analysis (RPCA) \cite{Candes 2011} and simultaneous decomposition and alignment of a stack of OCT images via linearized convex optimization, better performance is achieved in comparison to previous registration based denoising techniques. For more details on the method and results, the reader is referred to \cite{Peng 2012, Baghaie 2014b}

\begin{figure}
\begin{center}
\includegraphics[scale=.2]{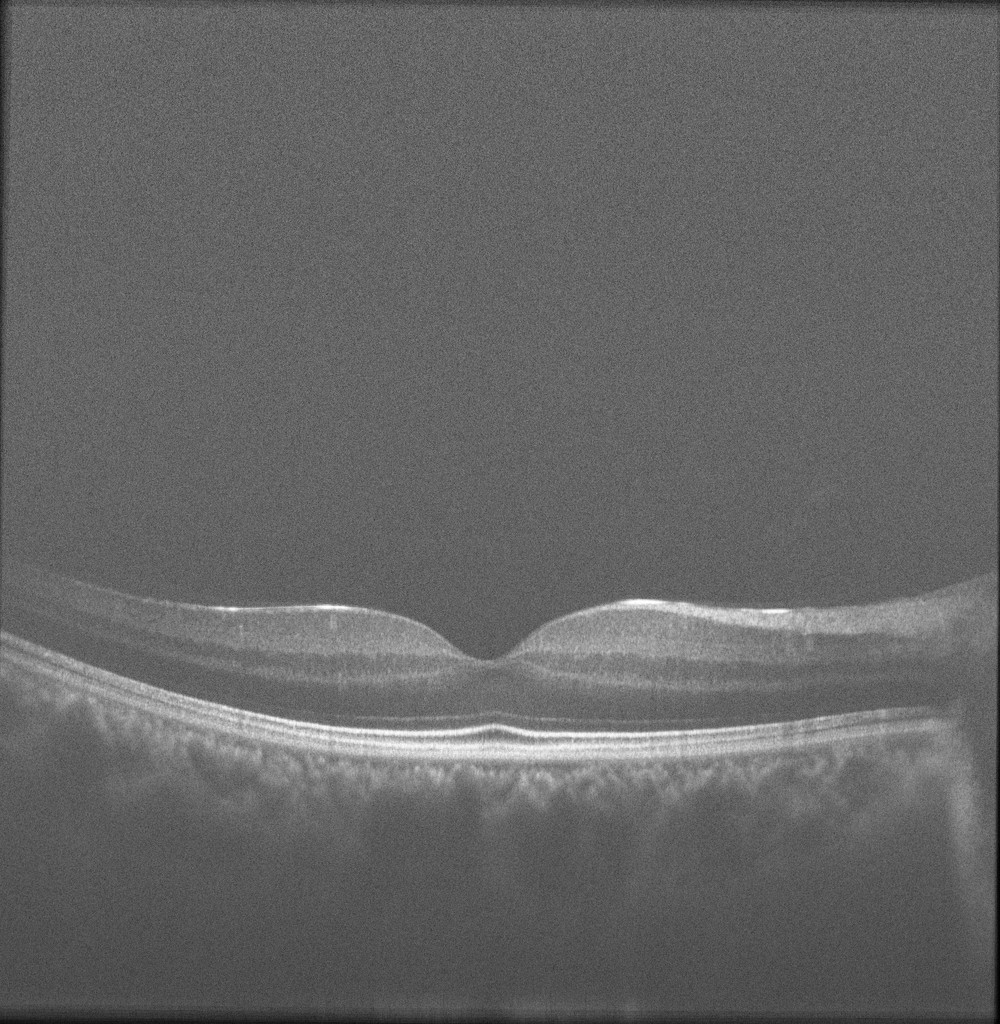}
\caption{Result of averaging affine registered cross-sections from close location in retina}
\end{center}
\end{figure}

Another application is in the registration of fundus images with \textit{en face} OCT images. This is very helpful to better correlate retinal features across different imaging modalities. In \cite{Li 2011} an algorithm is proposed for registering OCT fundus images with color fundus photographs. In this paper, blood vessel ridges are taken as features for registration. A similar approach is taken in \cite{Golabbakhsh 2013} too. Using the curvelet transform, the blood vessels are extracted for the two modalities and then image registration is done in two steps: 1) registration based on only scaling and translation and 2) registration based on a quadratic transformation. Fig. 9 shows different stages of the results of color fundus and \textit{en face} OCT image registration of the method proposed in \cite{Li 2011}.

\begin{figure}
\begin{center}
\includegraphics[scale=.3]{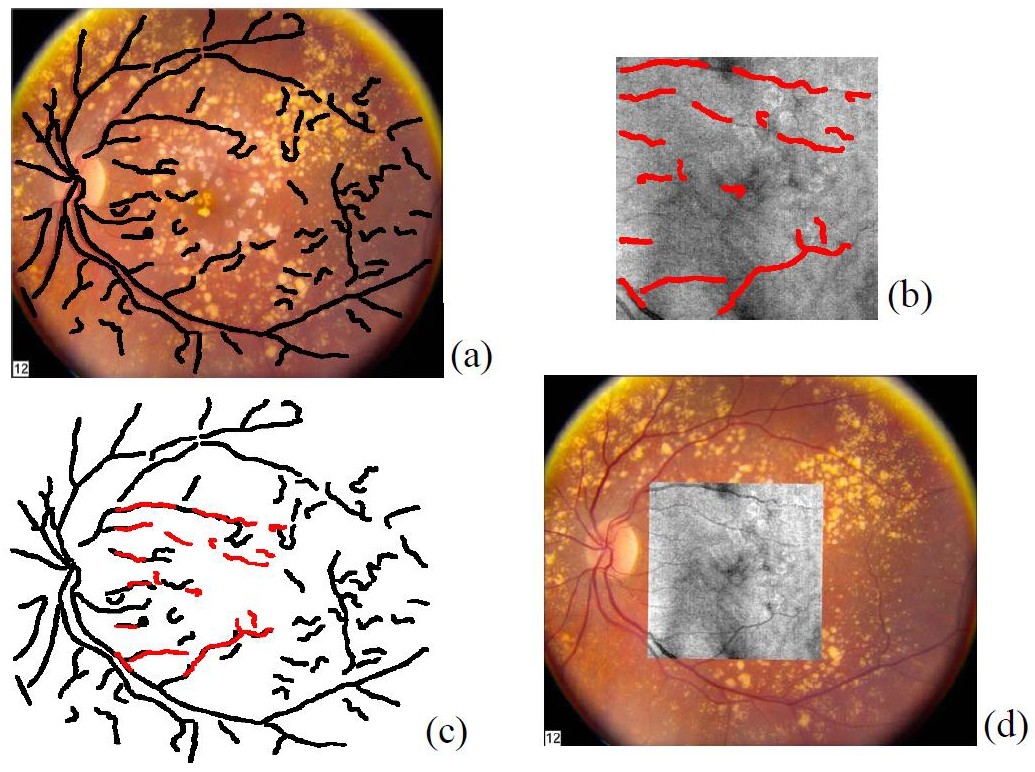}
\caption{Color fundus reference image superimposed by the corresponding blood vessel ridge image (a), \textit{en face} OCT image superimposed by the corresponding blood vessel ridge image (b), the registration result of blood vessel ridge image (c) and intensity images (d) \cite{Li 2011}}
\end{center}
\end{figure}

Motion correction of OCT images is another area in which image registration techniques are of high interest. There are several different involuntary eye movements that can happen during the fixation: tremor, drifts and micro-saccades \cite{Hart 1992}. One way to deal with this issue is a hardware solution which involves having eye tracking equipment to compensate the eye movement during image acquisition time. Usually a Scanning Laser Ophthalmoscopy (SLO) device is merged with the OCT for tracking the eye movements during imaging \cite{Pircher 2007}. Taking a software approach can be more general and applicable without the need for additional imaging equipment. In \cite{Ricco 2009} at first the vessels are detected and then using an elastic registration technique the tremor and drift motions are corrected. Micro-saccades are corrected in the next step by finding the horizontal shift at each pixel in the scan that best aligns the result of tremor and drift correction with the SLO image. In \cite{Xu 2012} a 3D aligning method for motion correction is proposed based on particle filtering. Fig. 10 shows one example of the use of the algorithm in \cite{Xu 2012} for correcting the motion in OCT images. Other techniques can be found in \cite{Kraus 2012, Kraus 2014} and references therein.

\begin{figure}
\begin{center}
\includegraphics[scale=.35]{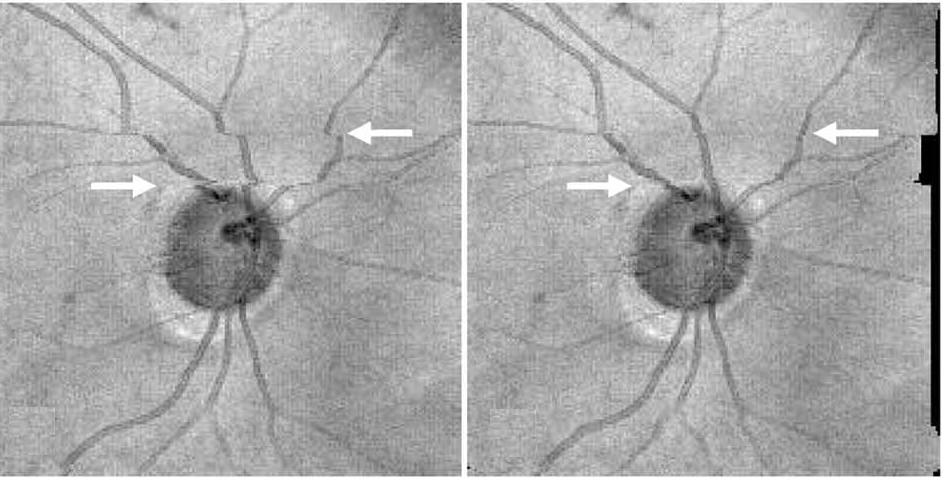}
\caption{OCT image with blood vessel discontinuity caused by eye movement (left) and corrected image (right) \cite{Xu 2012}}
\end{center}
\end{figure}

Image mosaicing is another example of using image registration methods in OCT image processing. OCT systems are capable of acquiring a huge amount of data in a very short period of time; but still the field of view is very limited. Stitching several volume data from a patient can improve the interpretation of data significantly. Usually in such methods, a set of overlapping OCT data is acquired and then stitched together using global/local image registration techniques. In \cite{Li 2011b} at first a set of 8 overlapping 3D OCT volume data over a wide area of retina is acquired. Then the OCT \textit{en face} fundus images are registered together using blood vessel ridges as features of interest. Finally the 3D OCT data sets are merged together using cross-correlation as a metric. A relatively similar approach is taken in \cite{Hendargo 2013} too for motion correction and 3D volume mosaicing of OCT images. Fig. 11 shows a color encoded depth image of retina with three main vessel layers. One of the very recent works in this area can be found in \cite{Lurie 2014} for bladder OCT images which is used with an integrated White Light Cystoscopy (WLC) system.

\begin{figure}
\begin{center}
\includegraphics[scale=.3]{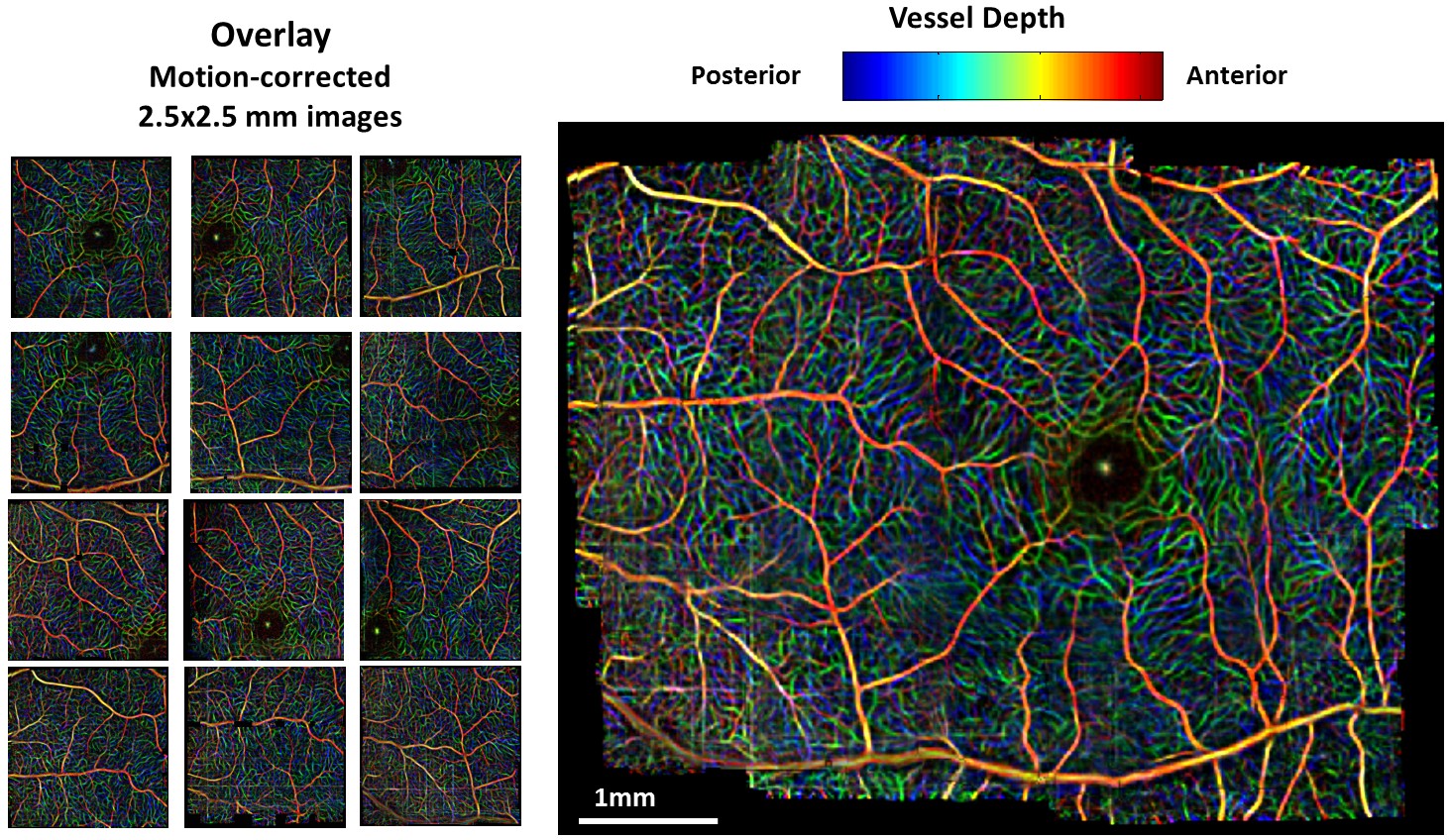}
\caption{Widefield mosaic of retinal layers displaying three main vessel layers. The initial images are on the left and the final result is on the right\cite{Hendargo 2013}}
\end{center}
\end{figure}

The use of image registration techniques in analysis and processing of OCT images has grown significantly in the past few years. Here only a few areas are mentioned and there are more to explore. 

\section{Conclusion}
Even though the techniques introduced here don not represent a comprehensive list of all of the approaches that are used for analysis of OCT images, the variety of the techniques gives us some pointers on the possible directions for further explorations. Especially with the emergence of parallel computing and GPU programming in the past few years, it is possible to have real time pre-processing and analysis systems to help with the increasing amount of data produced by OCT imaging systems. In this article, an overview of 3 major problems in OCT image analysis is provided: noise reduction, image segmentation and image registration, and several different techniques in each category are introduced. Starting from noise reduction, both hardware and software techniques are discussed to some extent. Moving to image segmentation, different segmentation techniques with a focus on retinal layer segmentation are introduced. Finally some of the image registration methods that are used for noise reduction, motion correction and image mosaicing of OCT images are reviewed. Overall, the use of image analysis techniques for OCT is a rapidly growing field and there remain many areas for investigation.

\end{document}